\documentclass[onecolumn,twoside,journal]{IEEEtran}
\usepackage{mathpple}
\usepackage{times}

\usepackage{amsmath}  
\usepackage{amssymb}  
\usepackage{mathrsfs} 

\usepackage{theorem}  
\usepackage{cite}     
\usepackage{comment}  
\usepackage{hyperref}

\usepackage{upref}
\usepackage{amsfonts}

\usepackage{verbatim}

\usepackage[dvipsnames,usenames]{color}

\usepackage{graphicx}
\usepackage{tikz}
\usetikzlibrary{calc,arrows.meta,positioning}
\usepackage{bm}

\usepackage{latexsym}
\usepackage[draft,ulem=normalem]{changes}
\usepackage{makecell}

\usepackage{enumitem}

\usepackage{algorithm}
\usepackage{algorithmic}

\allowdisplaybreaks

\newcommand{\be}[1]{\begin{equation}\label{#1}}
\newcommand{\ee}{\end{equation}}

\newcommand{\bc}{\begin{center}}
\newcommand{\ec}{\end{center}}

\renewcommand{\leq}{\leqslant}

\renewcommand{\geq}{\geqslant}

\newcommand{\Cref}[1]{Co\-rol\-la\-ry\,\ref{#1}}

\theoremstyle{plain} \theorembodyfont{\normalfont\slshape}

\newtheorem{thm}{Theorem$\!$}
\newenvironment{theorem}{\begin{thm}\hspace*{-1ex}{\bf.}}{\end{thm}}

\newtheorem{prop}[thm]{Proposition$\!$}

\newtheorem{lem}[thm]{Lemma$\!$}

\newtheorem{cor}[thm]{Corollary$\!$}

\newtheorem{defi}[thm]{Definition$\!$}

\newtheorem{cl}[thm]{Claim$\!$}

\theorembodyfont{\normalfont}

\newtheorem{exam}{Example$\!$}

\newtheorem{remrk}{Remark$\!$}

\newtheorem{const}{Construction$\!$}

\begin{document}

\title{Don't Fear the Bit Flips: Optimized Coding Strategies for Binary Classification}

\author{{\large Frederic~Sala, Shahroze~Kabir, Guy~Van~den~Broeck, and Lara~Dolecek} \\ University of California, Los Angeles \\ \texttt{ \{fredsala, shkabir\}@ucla.edu, guyvdb@cs.ucla.edu, dolecek@ee.ucla.edu}

}
\maketitle

\bibliographystyle{ieeetr}

\begin{abstract} 
After being trained, classifiers must often operate on data that has been corrupted by noise. In this paper, we consider the impact of such noise on the features of binary classifiers. Inspired by tools for classifier robustness, we introduce the same classification probability (SCP) to measure the resulting distortion on the classifier outputs. We introduce a low-complexity estimate of the SCP based on quantization and polynomial multiplication. We also study channel coding techniques based on replication error-correcting codes. In contrast to the traditional channel coding approach, where error-correction is meant to preserve the data and is agnostic to the application, our schemes specifically aim to maximize the SCP (equivalently minimizing the distortion of the classifier output) for the same redundancy overhead.
\end{abstract} 
\section{Introduction}
Noise is an enduring component of all computing and communication systems. Information is corrupted when transmitted over noisy channels \cite{Shannon}, stored in unreliable memories \cite{Dilillo}, or processed by noisy or low-quality circuits \cite{Mittal}. Moreover, techniques aimed at saving power or increasing efficiency can increase noise. For example, voltage scaling further increases the probability of data corruption \cite{Chandra}. The traditional approach to handling noisy storage media is to implement strategies to detect and correct errors. For example, many memories and drives implement error-correcting codes \cite{Sridharan}.

The typical policy of such systems is to ensure that information read back is the same as it was when written. Such a policy represents a very strong constraint; implementations tend to be expensive, with significant storage overhead dedicated to redundant data. In this paper, we ask whether we may relax this constraint in the context of feature data for classifiers. Our goal is to reduce the effect of noise on the output of the algorithm; in other words, we consider the distortion of the algorithm outputs, rather than the inputs.

\tikzstyle{place}=[circle,draw=black!100,fill=white!100,thick,inner sep=3.0pt,minimum size=1mm]
\usetikzlibrary{decorations.markings}
\tikzstyle{place2}=[rectangle,draw=black!100,fill=white!100,thick, inner sep=3.5pt]
\begin{figure}
	
	\centering
	\begin{tikzpicture}[inner sep=1mm]
	\node[place](C) at (0.0, 0.0) {$C$};
	\node[place](X1) at (-1.0,-1.0) {$X_1$};
	\node[place](X2) at (0.0,-1.0) {$X_2$};
	\node[place](XN) at (1.5,-1.0) {$X_n$};
	\node[place2,rotate=90](B1) at (-1.0,-2.5) {BSC$(\epsilon_1)$};
	\node[place2,rotate=90](B2) at (-0.0,-2.5) {BSC$(\epsilon_2)$};
	\node[place2,rotate=90](BN) at (1.5,-2.5) {BSC$(\epsilon_n)$};
	\node[place](Z1) at (-1.0,-4.0) {$X'_1$};
	\node[place](Z2) at (0.0,-4.0) {$X'_2$};
	\node[place](ZN) at (1.5,-4.0) {$X'_n$};
	\node at ($(X2)!.5!(XN)$) {\ldots};
	\node at ($(B2)!.5!(BN)$) {\ldots};
	\node at ($(Z2)!.5!(ZN)$) {\ldots};
	\node at (4.0, -0.6) {BSC$(\epsilon_i)$};

	\draw[->] (C)--(X1)  node[midway,above] {};
	\draw[->] (C)--(X2)  node[midway,above] {};
	\draw[->] (C)--(XN)  node[midway,above] {};
	\draw[->] (X1)--(B1)  node[midway,above] {};
	\draw[->] (X2)--(B2)  node[midway,above] {};
	\draw[->] (XN)--(BN)  node[midway,above] {};
	\draw[->] (B1)--(Z1)  node[midway,above] {};
	\draw[->] (B2)--(Z2)  node[midway,above] {};
	\draw[->] (BN)--(ZN)  node[midway,above] {};

	\node[black](I1) at (3.0, -1.5) {$1$};
	\node[black](Im1) at (3.0,-3.0) {$0$};
	
	\node[black](O1) at (5.0,-1.5) {$1$};
	\node[black](Om1) at (5.0,-3.0) {$0$};
	
	\draw[] (I1)--(O1)  node[midway,above] {$1-\epsilon_i$};
	\draw[] (I1)--(Om1) node[midway, above] {$\epsilon_i$\quad\quad\quad\quad};
	
	\draw[] (Im1)--(O1) node[midway,below] {$\epsilon_i$};
	\draw[] (Im1)--(Om1) node[midway,below] {$1-\epsilon_i$};
	\draw[] (2.8,-1.0) rectangle (5.2,-3.5);

	\end{tikzpicture}
	\caption{Noisy classifier process model. The noiseless classification uses the features $X_1, X_2, \ldots, X_n$. Noisy feature $X'_i$ is produced from $X_i$ with the binary symmetric channel BSC$(\epsilon_i)$: $X'_i$ is equal to $X_i$ with probability $1-\epsilon_i$ and equal to $\bar{X_i}$ with probability $\epsilon_i$. Classification is performed on the noisy features $X_1', \ldots, X'_n$, without observing the true values $X_1, \ldots, X_n$.}
	\label{fig:setup}
\end{figure}
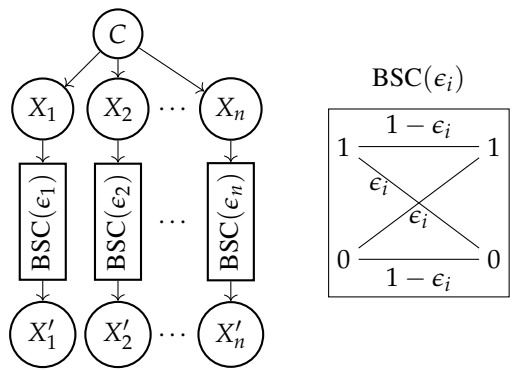

The basic setup is shown in Figure~\ref{fig:setup}. We have a na\"{\i}ve Bayes classifier with class variable $C$ and binary features $X_1,X_2, \ldots, X_n$. Learning is performed on noiseless training data. Afterwards, the classifier operates on noisy data; rather than the true features $X_1, \ldots, X_n$, only the corrupted features $X'_1, X'_2, \ldots, X'_n$ are used. Feature $X_i$'s value is flipped from 0 to 1 or from 1 to 0 with probability $\epsilon_i$. A fundamental question is to determine the \emph{same classification probability} (SCP), that is, the probability that the classifier output is the same for $X_1, \ldots, X_n$ and for $X'_1, \ldots, X'_n$.

Features can be protected by using an error-correction strategy. A set of redundant bits is computed when the data is written; before the data is used, both the data and redundant bits are decoded, with the goal of correcting some of the errors. This procedure is known as \emph{channel coding}. The traditional approach is to protect all bits equally, uniformly reducing the $\epsilon_i$s for all $n$ features. As we will see, this is not an efficient approach. Some features are more valuable and more worthy of protection than others. In fact, given a budget of redundancy, our major goal is an optimal resulting set of $\epsilon_1, \ldots, \epsilon_n$ to minimize the classifier distortion. 

We illustrate this idea with a simple example. Take a classifier with $n=2$ features $X_1, X_2$, noisy versions $X'_1, X'_2$, and uniform prior. The noise parameter for both binary symmetric channels (BSCs) is $\epsilon = 0.1$ by default. Let $p_i(x_j)$ denote the conditional probability $p(x_j = 0|C = i)$. We show how the SCP varies for different values of the $p_i(x_j)$'s and noise parameters. We also demonstrate how protecting features uniformly or non-uniformly affects the SCP. ``Protection'' refers to an error-correction scheme that effectively reduces the $\epsilon$ noise parameter for a particular feature. Here, we allocate 4 additional redundancy bits for protection; uniform protection gives 2 such bits to each feature, while protection on $X_i$ alone grants all 4 bits to $X_i$. (More precise definitions for this terminology are provided later on in the paper; the result of protection is a reduction of the $\epsilon_i$ noise parameter for feature $i$.)

\begin{table}[t]
\small
\addtolength{\tabcolsep}{-4pt}
\centering
\caption{SCPs for various redundancy allocations}
\label{tb:ex1}
{\begin{tabular}{cccc|cccc}
\multicolumn{4}{c|}{Parameters} & \multicolumn{4}{|c}{SCP by protection strategy}\\
$p_0(x_1)$ & $p_0(x_2)$ & $p_1(x_1)$ & $p_1(x_2)$ & Uniform & None & All~$x_1$ & All~$x_2$ \\ \hline
0.9    & 0.89   & 0.1    & 0.11   & 0.963                & 0.900         & \textbf{0.991}      & 0.900      \\ 
0.8    & 0.81   & 0.3    & 0.3   & \textbf{0.964}                & 0.905        & 0.946     &0.946     
\end{tabular}}
\end{table}

In the first row of Table~\ref{tb:ex1}, $X_1$ contains more information about $C$ compared to $X_2$. Observe that allocating all bits to $X_1$ yields a better SCP than equal protection or protection on $X_2$. Conversely, if we look at the last row we see that even though $X_2$ contains more information about the class variable, the SCP is maximized by a uniform allocation of redundancy bits. Thus we cannot simply examine the conditional probabilities to decide how to allocate redundancy. Such results motivate us to seek an informed redundancy allocation strategy.

%


Our contributions are 
\begin{itemize}
\item A framework for measuring the impact of noise by the \emph{same-classification probability} (SCP),
\item A low-complexity approximation for the SCP based on quantization and polynomial multiplication,
\item A study of coding strategies based on allocating redundant bits in a way that minimizes the SCP (and thus the distortion on the classifier output). We show that, surprisingly, the empirical SDP is non-monotonic with respect to error protection. Afterwards, we give an optimization for optimal redundancy allocation based on a greedy approach.
\end{itemize}

\section{Prior Work}
Channel coding for data protection is a vast field. Error-correction has been proposed for disk drives \cite{Riggle}, write-once memories \cite{Rivest}, and for RAID architectures \cite{Blaum}. More recently, techniques have been developed to take advantage of the specific properties of modern non-volatile memories such as flash \cite{Cassuto,Dolecek}. Distributed storage systems can be protected through replication or erasure coding \cite{Weatherspoon}, \cite{Dimakis}. Solid state drives can also benefit from sophisticated error-correction \cite{KZhao}. A common aspect of such research, in contrast to our work, is that the goal is to preserve the data being stored without considering the application. That is, error-correction is found in a different abstraction layer.

The research in \cite{Mazooji16} introduces an informed channel coding scheme for linear regression. Here, the error model was also bit flips applied to a finite-width binary representations of integers. The goal was to minimize the distortion on the algorithm outputs. An approximation scheme was introduced by considering the contribution to the distortion by a flip in a single bit independently of the others. 

There are a number of papers that have examined issues related to learning algorithms dealing with uncertainty. Recent work on information dropout, a technique introduced in \cite{Srivastava} to prevent neural networks from overfitting, has considered adding noise to the activations of deep neural networks \cite{Achille}; unlike our work, here noise is used during learning. 

The robustness of various algorithms under noise exposure has been considered in a series of papers. \cite{Kala} performed an experimental study of algorithms when data sets were corrupted by synthetic noise. Deleted features were tackled in \cite{Dekel} and \cite{Globe}. \cite{Dekel}, which is perhaps the most closely related work to the present paper, also considers corrupted features, and introduces two techniques to tackle noise, one based on linear programming, and the other using an online batching scheme. However, unlike our work, this paper considers only adversarial (not probabilistic) noise and does not propose optimized channel coding. \cite{Globe} uses a game-theoretic approach to avoid over-reliance on a feature that could be deleted. Another work tackling the case of missing data in Bayes classifiers is \cite{Ramoni}.

Noise and coding are often important issues in distributed learning. For example, \cite{Zhang} introduced optimality guarantees for distributed estimators in a setting where nodes can be isolated, but limited communication between them is allowed. \cite{Han} considered problems including hypothesis testing and parameter estimation in the case of multiterminal algorithms where each terminal has a data compression constraint.

A related area is known as value of information~\cite{Heckerman93,bilgic11}. The idea is to maximize an expected reward on some observed features. One example is to observe those features that maximize the information gain \cite{krause2009}.

Decision robustness in the case of hidden variables can be measured by the \emph{same decision probability} (SDP) \cite{darwiche12,chen2014}. SDP has been applied to the evaluation of adaptive testing~\cite{chen2015computer} and of mammography-based diagnosis~\cite{gimenez2014novel} The same classification probability we employ in this paper bears some similarity to the SDP. The problem of handling label noise is considered in \cite{Frenay}. Label noise in crowdsourcing is tackled through coding-theoretic means in \cite{Vempaty}. Additionally, our work can be viewed as a form of feature selection \cite{Guyon}, although we do not remove features, but rather allow them to be noisier than certain more critical features.

\section{Noise in Binary Classification}
We use the standard notation of upper-case letters for random variables and lower-case letters for their instantiations. Boldface denotes vectors of variables. 

Throughout this paper we consider binary classification problems with binary features. For some test point ${\bf x} =  (x_1, x_2, \ldots, x_n)$ with $n$ binary features, we denote the conditional probabilities by $\alpha_i = p(x_i|C=0)$ and $\beta_i = p(x_i|C=1)$ for $1 \leq i \leq n$. Throughout this paper, we employ a na\"{\i}ve Bayes classifier such that ${\bf x}$ is classified to 
\[c({\bf x}) = \text{arg} \max \left\{p(C=0)\prod_{i=1}^n \alpha_i, \text{ } p(C=1) \prod_{i=1}^n \beta_i.\right\}\]

Taking logarithms, $\bf x$ is classified according to to the value of 
\begin{equation} \label{class_eqn}
c({\bf x}) = sgn \left(\log\frac{p(C=0)}{p(C=1)} + \sum_{i=1}^n \log(\alpha_i/\beta_i) \right).
\end{equation}

\subsection{Impact of Noise}



In this work, we employ a simple noise model: we define an \emph{noise parameter vector} $\bm{\epsilon} = (\epsilon_1, \epsilon_2, \ldots, \epsilon_n)$ with $0 \leq \epsilon_i < 1/2$ for $1 \leq i \leq n$. The binary feature $X_i$ is flipped to its opposite value $\bar{X_i}$ with probability $\epsilon_i$ and stays unchanged with probability $1-\epsilon_i$. Using information theory terminology, we view this operation as placing a binary symmetric channel on each of the features, as shown in Figure~\ref{fig:setup}. We express the resulting \emph{error vector} as ${\bf E} = (E_1, E_2, \ldots, E_n)$, where $E_i = 1$ if an error has occurred for feature $i$ and 0 otherwise. As a special case, we consider the vector $\bm{\epsilon} = (\epsilon, \epsilon, \ldots, \epsilon)$, where the error probability is identical on all features. In this setting, the probability of some error instantiation $\bf e$ is a function of its Hamming weight $wt({\bf e})$: 
\[Pr({\bf e}) = \epsilon^{wt({\bf e})} (1-\epsilon)^{n-wt({\bf e})}.\]

\subsection{Same Classification Probability}
Now we examine the effect of the feature bit errors on the output of the classification algorithm. Our goal is to determine when noise does not impact the algorithm output. That is, we wish to compute the probability that the noiseless point $\bf X$ and the noisy version ${\bf X}({\bf E}) = (x'_1, \ldots, x'_n)$ have the same classification:
\begin{equation}\label{eq:scp_real}
\text{SCP}({\bf X}, {\bf E}) = Pr(c({\bf X}) = c({\bf X}({\bf E}))).
\end{equation}
We call this quantity the (real) \emph{same classification probability}. We are mainly concerned with an empirical version of the SCP for a feature realization ${\bf x}$,
\begin{equation}\label{eq:scp}
\text{SCP}({\bf x}, {\bf E}) = \sum_{{\bf e} \in \{0,1\}^n} Pr(c({\bf x}) = c({\bf x}({\bf e}))) Pr({\bf e}).
\end{equation}
The SCP resembles the same decision probability (SDP) introduced in \cite{darwiche12}, defined as  the probability of producing a decision (using evidence ${\bf e}$) confirmed by $Pr(d|{\bf e} \geq T)$ as when some hidden variable $\bf H$ is revealed, $\sum_{\bf h} Pr(d | {\bf e}, {\bf h}) Pr({\bf h}|{\bf e})$.  A key difference from SDP is that SCP considers two separate distributions: the augmented distribution that captures the data generation process (Figure~\ref{fig:setup}), and the original na\"{\i}ve Bayes network where classification is performed. Instead, SDP is calculated a single Bayesian classifier that is assumed to capture both processes. 

If $e_j = 1$ and thus there is a feature bit flip at position $j$, the corresponding term in \eqref{class_eqn} changes as \[\log \left(\frac{\alpha_j}{\beta_j}\right) \rightarrow \log \left( \frac{1-\alpha_j}{1-\beta_j}\right).\]

We denote $A_j := \log \left(\frac{\alpha_j}{\beta_j}\right) $ and $B_j := \log \left( \frac{1-\alpha_j}{1-\beta_j}\right)$. Let us write $D_j$ for the difference 
\[D_j = B_j - A_j = \log \left( \frac{1-\alpha_j}{1-\beta_j}\right) - \log \left(\frac{\alpha_j}{\beta_j}\right).\] An error $e_j=1$ replaces $A_j$ in \eqref{class_eqn} with $B_j$; equivalently, $D_j$ is added to this sum. Therefore, for some error vector $\bf e$ with $\ell$ errors given by the components $e_{i_1}, e_{i_2}, \ldots, e_{i_{\ell}} = 1$, the noisy classification is given by 
\begin{equation} \label{class_eqn_noisy}
c({\bf x}({\bf e})) = sgn \left(\log\frac{p(c=0)}{p(c=1)} + \sum_{i=1}^n A_i + \sum_{j=1}^{\ell} D_{i_{j}} \right).
\end{equation}
Producing the same classification requires $c({\bf x}({\bf e})) = c({\bf x})$, or, equivalently, 
\begin{align}
\label{eq:sdp2}
&sgn \left(\log\frac{p(c=0)}{p(c=1)} + \sum_{i=1}^n \log(\alpha_i/\beta_i) \right) \nonumber \\
&\quad = sgn \left(\log\frac{p(c=0)}{p(c=1)} + \sum_{i=1}^n A_i + \sum_{j=1}^{\ell} D_{i_{j}} \right). 
\end{align}

Without loss of generality, fix the noiseless classifier output to $c=0$, so that the sign function value is positive\footnote{If the noiseless classifier output is $c=1$, the outcome is that we seek subsets whose sum is less than the target value $T$ rather than greater. All of the arguments we present in this paper are unchanged. For simplicity of notation, we do not add the case to our setup.}. Next, write \[T := -\left(\log\frac{p(c=0)}{p(c=1)} + \sum_{i=1}^n A_i\right).\] Here, $T$ represents a target value. Then, the equality \eqref{eq:sdp2} is equivalent to 
\begin{equation} \label{subset_sum}
\sum_{j=1}^{\ell} D_{i_j} > T
\end{equation}

Recall that the $D_{i_j}$ correspond to all the components in ${\bf e}$ that are equal to 1. Then, to sum over all ${\bf e}$ in \eqref{eq:scp}, we must determine all the subsets of $\{D_1, D_2, \ldots, D_n\}$ with sum greater than the target $T$. A subset of size $\ell$ corresponds to an error vector with $\ell$ feature bit errors that does not change the classification relative to the noiseless version. Such a subset contributes a term (given by \eqref{eq:scp}) to the same classification probability. 

{\bf Example:} Consider a model with $n=3$ binary features and equal error probability $\bm{\epsilon} = (\epsilon, \epsilon, \epsilon)$. Take $D_1+D_2 > T$. This implies that the error vector ${\bf e} = (1,1,0)$ does not change the classification output, contributing a probability term of $\epsilon^2 (1-\epsilon)$ to the SCP.

\section{SCP Approximation}

In this section, we consider issues surrounding the computation of the SCP. Our main result is an efficient SCP approximation.

\subsection{Exact Empirical SCP}
The problem described by \eqref{subset_sum} is a variant of the subset sum problem with real numbers. The na\"{\i}ve approach is to enumerate the $2^n$ subsets of $\{D_1, D_2, \ldots, D_n\}$ and compute each of their sums. This task is not tractable for large $n$.  

A compact representation of the SCP is the function
\begin{align*}
G(z) &= \prod_{i=1}^n ((1-\epsilon_i)+\epsilon_iz^{D_i}) 
\end{align*}
We define $[>T]G(z)$ to be the sum of the coefficients of terms in $G(z)$ with exponent larger than $T$. It is easy to see that \[\text{SCP}({\bf x}, {\bf E}) = [>T]G(z).\] Any subset $\{D_{i_1}, D_{i_2}, \ldots, D_{i_j}\}$ with sum greater than $T$ produces a term in $G(z)$ with exponent $D_{i_1} + \ldots + D_{i_j} > T$ and coefficient \[\epsilon_{i_1} \epsilon_{i_2} \cdots \epsilon_{i_j} \prod_{i \neq i_1, \ldots, i_j} (1-\epsilon_i).\] The sum of all such coefficients is indeed the SCP. This may lead us to attempt to expand the function $G(z)$ and examine the resulting coefficients. However, the fact that the exponents are real-valued prevents us from doing so efficiently. 

\subsection{Quantization}
Observe that if the exponents of terms in $G(z)$ were non-negative integers, $G(z)$ would be a polynomial, allowing for fast multiplication. In fact, $G(z)$ would become a \emph{generating function} \cite{gfref}. Inspired by this notion, we introduce the following quantization scheme.

The key idea is to quantize $D_i$ into $k$ buckets, for $k$ a constant. By performing the quantization in a clever way, we induce a structure that enables us to approximate the SCP in no more than $O(n^2 \log n)$ operations. The approximation is described in Algorithm~\ref{alg:sdpapp}; the concept is detailed below.


Our quantization scheme only relies on the minimal and maximal values of $\{D_1, \ldots, D_n\}$. Let $D_{min} = \min \{D_1, \ldots, D_n \}$ and $D_{max} = \max \{D_1, \ldots, D_n\}$. Let $D_I = D_{max}-D_{min}$. Consider the family of $k$ intervals

\begin{align}
\label{eq:intervals}
& \left[D_{min}-\frac{D_I}{2(k-1)},D_{min}+\frac{D_I}{2(k-1)} \right), \left[D_{min}+\frac{D_I}{2(k-1)},D_{min}+\frac{3D_I}{2(k-1)}  \right), \ldots, \nonumber \\ 
&\qquad \left[D_{min}+\frac{(2k-5)D_I}{2(k-1)},D_{min}+\frac{(2k-3)D_I}{2(k-1)}  \right), \left[D_{min}+\frac{(2k-3)D_I}{2(k-1)} , D_{max}+\frac{(2k-1)D_I}{2(k-1)} \right).
 \end{align}

Note that each bucket has width $D_i/(k-1)$. For compactness, we also write the $i$th interval as $[E_i^{\text{begin}}, E_i^{\text{end}})$. 

We quantize any value of $D_j$ that falls into a bucket with the midpoint of the bucket interval. We compute $S_i$, the number of $D_j$ in each bucket:

\begin{align*} S_i = &\left|\left\{ D_j : D_j \in \left[D_{min}+\frac{(2i-3)D_I}{2(k-1)}, D_{min}+\frac{(2i-1)D_I}{2(k-1)}  \right), 1 \leq j \leq n \right\} \right|,\end{align*}
for $1 \leq i \leq k$. In addition, we re-label the noise parameters such that $\epsilon_{i1}, \epsilon_{i2}, \ldots, \epsilon_{iS_i}$ are the noise parameters corresponding to the features $D_j$ that fall into the $i$th bucket.

Recall that we seek to determine the subsets of $\{D_1, D_2, \ldots, D_n\}$ that have sum greater than $T$. We approximate any $D_j$ that falls into the $i$th interval with the midpoint of that interval:
\begin{equation}
\label{eq:midpoint}
D_{min} + \frac{i-1}{k-1}D_I.
\end{equation}
Additionally, observe that $D_{min}$ and $D_{max}$ falls are represented by their own values.

We approximate the SCP by computing the SCP on the quantized versions of the $D_i$; we show that this can be performed efficiently by Algorithm~\ref{alg:sdpapp}.

\begin{algorithm}[tb]
   \caption{SCP Approximation}
   \label{alg:sdpapp}
\begin{algorithmic}
   \STATE {\bfseries Input:} Difference terms $D_1, D_2, \ldots, D_n$, Error probabilities $\epsilon_1, \epsilon_2, \ldots, \epsilon_n$, Target $T$, Number of buckets in quantization scheme $k$
   \STATE {\bfseries Output:} SCP approximation $\text{SCP}_{app}$
   \STATE {\bfseries Initialize} $S_1, \ldots, S_k$ to 0, $\text{SCP}_{app}$ to 0
   \FOR{$j=1$ {\bfseries to} $n$}
   \IF{$D_j \in [E_i^{\text{begin}}, E_i^{\text{end}})$} 
   \STATE $S_i \leftarrow S_i + 1$
   \STATE $\epsilon_{iS_i} \leftarrow \epsilon_j$
   \ENDIF
   \ENDFOR
   \STATE {\bf expand} $G[i,j] = \prod_{i=1}^k \prod_{j=1}^{S_i} ((1-\epsilon_{ij}) + \epsilon_{ij} yz^i)$
   \FOR{$i=0$ {\bfseries to} $n$}
   \STATE $T' \leftarrow \frac{k-1}{D_I}\left(T-i \left(D_{min}-\frac{D_I}{k-1}\right)\right)$
   \FOR{$j\geq 0$}
   \IF{$G[i,j] > T'$} 
   \STATE $\text{SCP}_{app}$ $\leftarrow$ $\text{SCP}_{app}$  + $G[i,j]$
   \ENDIF
   \ENDFOR
   \ENDFOR
\end{algorithmic}
\end{algorithm}

Consider the multivariate generating function 
\begin{align}
\label{eq:genfun}
G(y,z) = \prod_{i=1}^k \prod_{j=1}^{S_i} ((1-\epsilon_{ij}) + \epsilon_{ij} yz^i) . 
\end{align} 
We write $[\ell, > R]G(y,z)$ for the sum of all coefficients of terms with exponent $\ell$ in $y$ and exponent greater than $R$ in $z$. Note that in the case where $\bm{\epsilon} = (\epsilon, \ldots, \epsilon)$, $G(y,z)$ reduces to 
\[G(y,z) = \prod_{i=1}^k ((1-\epsilon) + \epsilon yz^i)^{S_i}.\] 

We show that the appropriate coefficients in the generating function yield the SCP for the quantized versions of the $D_i$:
\begin{theorem}
\label{thm:quant}
Under the quantization scheme given by Algorithm~\ref{alg:sdpapp}, the resulting SCP is given by 
\[\text{SCP}_{app}({\bf x},{\bf E}) = \sum_{\ell=0}^n [\ell, > T'(\ell)]G(y,z),\]
where \[T'(\ell) :=\frac{k-1}{D_I}\left(T-\ell \left(D_{min}-\frac{D_I}{k-1}\right)\right).\]
\end{theorem}
\begin{IEEEproof}
Consider some subset with size $\ell$ and sum greater than the target $T$ in the quantization scheme given by \eqref{eq:intervals}. We distribute our $\ell$ choices of $D_j$ into the $k$ buckets, where we may take up to $S_i$ values from bucket $i$. Let $(a_1, a_2, \ldots, a_k)$ be such a choice; $a_i$, for $0 \leq a_i \leq S_i$ represents the number of $D_j$ falling into the $i$th bucket. A subset of size $\ell$ requires $a_1 + a_2 + \ldots + a_k = \ell$. The total sum for the subset is given by 
\[a_1D_{min} + a_2\left(D_{min} + \frac{D_I}{k-1}\right) + \ldots + a_kD_{max}.\]
Here we used the fact that terms in each bucket are quantized to the midpoint of that bucket (i.e., expression \eqref{eq:midpoint}). 


Since the sum of the subset is greater than $T$, we may write

\begin{align*}
&(a_1 + a_2 + \ldots + a_k)\left(D_{min}-\frac{D_I}{k-1}\right) + \left(a_1\frac{D_I}{k-1} + a_2\frac{2D_I}{k-1} + \ldots + a_k\frac{kD_I}{k-1}\right) > T \\
&\implies \ell \left(D_{min}-\frac{D_I}{k-1}\right) + \frac{D_I}{k-1} \left(a_1 + 2a_2 + \ldots + ka_k \right)  > T \\
&\implies a_1 + 2a_2 + \ldots + ka_k >  \frac{k-1}{D_I}\left(T-\ell \left(D_{min}-\frac{D_I}{k-1}\right)\right) := T'.
\end{align*}

Now we relate the generating function given in \eqref{eq:genfun} to the equation $a_1 + 2a_2 + \ldots + ka_k > T'$. The generating function is the product of $n$ binomials. To generate a term in the expansion, we select from each binomial either $1-\epsilon_{ij}$ or $\epsilon_{ij}yz^{i}$. The subsets of size $\ell$ satisfying $a_1 + 2a_2 + \ldots + ka_k > T'$ correspond to choosing $a_1$ terms $\epsilon_{1j}yz$, $a_2$ terms $\epsilon_{2j}yz^2$, and so on. The products of such terms have degree $\ell$ in $y$ and degree $a_1 + 2a_2 + \ldots + ka_k > T'$ in $z$. Considering the subsets of size $0,1,\ldots, n$, we indeed have that $\text{SCP}_{app}({\bf x},{\bf E}) = \sum_{\ell=0}^n [\ell, > T']G(y,z).$
\end{IEEEproof}

\subsection{Computation}
To compute $\text{SCP}_{app}({\bf x},{\bf E})$, we must perform the expansion of the generating function $G(y,z) = \prod_{i=1}^k \prod_{j=1}^{S_i} ((1-\epsilon_{ij}) + \epsilon_{ij} yz^i)$ and examine the coefficients. Polynomial multiplication is equivalent to convolution (of vectors or arrays, depending on whether the polynomials are univariate or multivariate); this effort can be sped up by using the Fast Fourier Transform (fft) \cite{fftref}. In the Fourier domain, convolution is equivalent to multiplication. That is, for two polynomials $q(x), r(x)$, \[q(x)r(x) = \text{ifft}(\text{fft}(q) \times \text{fft}(r)),\]
where the multiplication on the right side is performed term by term and ifft denotes the inverse transform. If the maximum degree of $q(x)$ and $r(x)$ is $d$, the number of operations required is $O(d \log d)$.

First, we expand the terms $\prod_{j=1}^{S_i} ((1-\epsilon_{ij}) + \epsilon_{ij} yz^i)$. Replace $yz^i$ by $v$; then, we are multiplying $S_i$ polynomials of degree 1. This can be performed in $O(S_i \log \log S_i)$ operations. This fact is not difficult to check: our $S_i$ monomials can be paired up and the pairs multiplied; the $S_i/2$ resulting degree-2 polynomials are paired up, and so on. The total time is $O(S_1 \log \log S_1) + O(S_2 \log \log S_2) + \ldots + O(S_k \log \log S_k) \leq O(n \log \log n)$. We note that each expansion can be performed in parallel.

Note that in the case of identical $\epsilon$, the polynomial to be expanded is $((1-\epsilon) + \epsilon yz^i)^{S_i}$, and the expansion can be performed in $O(S_i)$ operations with the binomial theorem.


Next, each of the resulting expansions must be multiplied. The expansion of $\prod_{j=1}^{S_i} ((1-\epsilon_{ij}) + \epsilon_{ij} yz^i)$ can be represented as a two-dimensional array, where rows represent the exponent of $y$ and columns the exponent of $z$. In our case, we have $k$ arrays. The maximal degree for $y$ is $n$, while the maximal degree for $z$ is $kn$, so that the product of all of the arrays (and thus any intermediate multiplication) has size at most $n \times kn$. Each multiplication has cost at most $O(kn^2 \log(kn))$; the total cost is thus no more than $O(k^2n^2 \log(kn))$. (Again, it is possible to speed to perform the multiplications in parallel for speedup.) 

\subsection{Approximation Quality}
\begin{figure}
\centering
\includegraphics[width=0.6\columnwidth]{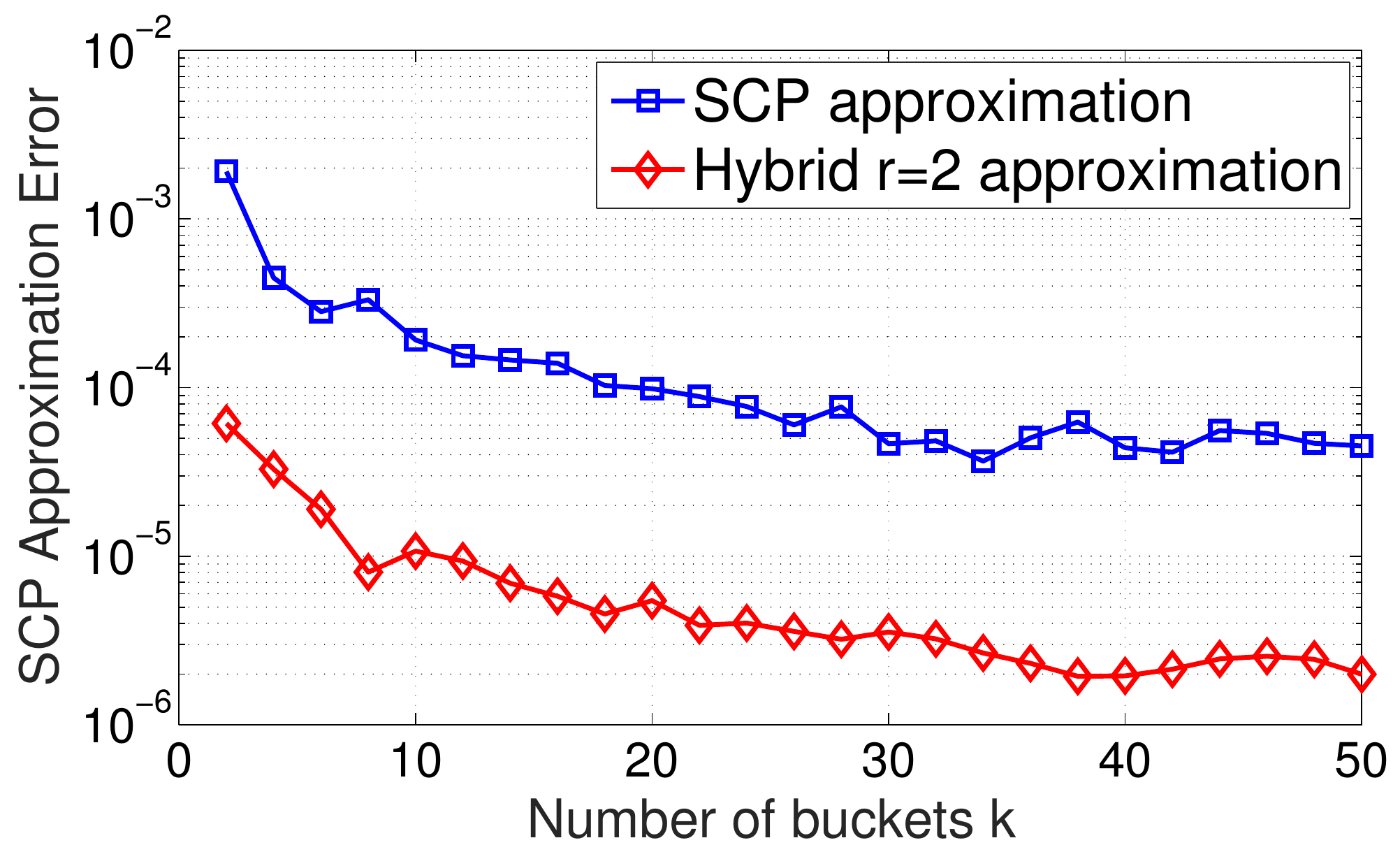}
\caption{SCP approximation error (against the real SCP) for the two approximation schemes. The parameters here are $\epsilon=10^{-2}$ and $n=16$. 50 data points were selected randomly; the true SCP was computed and the average absolute error was measured for various $k$ in the two schemes.}
\label{fig:app}
\end{figure}

Next we comment on the quality of the approximation. Clearly, the quality is a function of the number of buckets $k$; the larger the $k$, the finer the approximation. We note, however, that the approximation error is not a monotonic function, since there are edge effects for different $k$, as can be seen in Figure~\ref{fig:app}. 

We introduce two improvements to optimize the worst-case error of the SCP approximation. We describe these errors as a function of the noise parameter $\bm{\epsilon}$, and for simplicity take the case where all the components equal to $\epsilon$. The first improvement comes at no extra cost and reduces the worst-case error to $O(\epsilon^2)$, while the second improvement requires $O(n^r)$ time but reduces the worst-case error to $O(\epsilon^{r+1})$. We adopt the first improvement for our experiments.

{\bf Improvement 1}. For the first improvement, consider the intervals defined in \eqref{eq:intervals}. If our target value $T$ lines somewhere in $[D_{min}, D_{max}]$, we can translate (shift) all of our intervals over by some constant $T_{shift}$ that is no larger than half an interval width, $D_I/2(k-1)$, such that $T$ is now precisely the edge between two intervals. Then, any $D_i < T$ is mapped to a bucket center smaller than $T$, while any $D_i \geq T$ is mapped to a bucket center larger than $T$. We conclude that any single-element subset of $\{D_1, \ldots, D_n\}$ that contributes to the SCP also contributes to the SCP approximation; thus, all SCP approximation errors must be for two- or more element subsets, reducing this error to $O(\epsilon^2)$.

{\bf Improvement 2}. We refer to the second improvement as a \emph{hybrid SCP approximation}. The idea here is simple: we use the approximation only for the subsets of size greater than $r$ for some constant $2 \leq r < n$. For those subsets that are of size $r$ or smaller, we check the sums using the actual values of $D_i$, requiring at most $O(n^r)$ time. This strategy is most suitable for very small values of $r$, such as $r=2$. 

{\bf Example}. An evaluation was performed on the small data set house-votes-84 from the UCI repository \cite{Lichman}. This set has $n=16$ binary features corresponding to `yes' or `no' votes on various congressional proposals; the binary class represents democrat or republican. The small size of $n$ enables us to compute the real SCP value (though much slower than the approximate SCP). Figure~\ref{fig:app} shows the error versus the true SCP for the two approximation schemes and various $k$ averaged over $50$ random test points; the noise parameter was $\epsilon = 10^{-2}$. 






\section{Optimized Channel Coding}
Thus far we have only considered the impact of noise without taking any action to protect the algorithm output from distortion. In this section, we consider a simple but effective form of error correction. Our goal is to tailor this strategy to the characteristics of the features in order to minimize the classifier distortion. This is a departure from traditional error protection, which is typically present in a different abstraction layer and is agnostic to the algorithm.


Consider storing multiple copies of a certain feature bit; these copies are produced when the data is written and before the addition of noise. Afterwards, when the bit is ready to be used, the repeated values of the bit (some of which may now be corrupted) are subject to a majority vote (equivalent to maximum-likelihood decoding) and the decoded value is placed back into the algorithm.

What is the effect of repetition on the error probability $\epsilon$? If the $j$th feature $x_j$, corrupted by a bit flip with probability $\epsilon$, is repeated $2r+1$ times for some $r \geq 0$, the probability $\epsilon^{(r)}$ that the feature is decoded incorrectly is equal to the probability that the majority of votes are corrupted:
\[\epsilon^{(r)} = \sum_{i=r+1}^{2r+1} \binom{2r+1}{i} \epsilon^{i} (1-\epsilon)^{2r+1-i}.\]

For example, if $r=1$, the smallest meaningful repetition, $\epsilon^{(1)} = \epsilon^3 + 3\epsilon^2(1-\epsilon).$ Note that we always add pairs of repetitions, since an even number of repeated bits can produce ties. In general, employing $r$ pairs of repeated bits drives the error probability $\epsilon$ to $O(\epsilon^{1+r})$.

\subsection{Monotonicity}
Intuitively, we expect that reducing the error probability on features will increase the SCP. Clearly, reducing all of the $\epsilon_i$ to 0 yields an SCP of 1, as the only non-zero term in the SCP is $(1-\epsilon_1)(1-\epsilon_2)\times\ldots\times(1-\epsilon_n) = 1$. 

Nevertheless, the SCP does not necessarily increase when we reduce noise parameters $\epsilon$. The following result is applicable to both the SCP and the SCP approximation:

\begin{algorithm}[tb]
   \caption{Channel Code Optimization}
   \label{alg:code}
\begin{algorithmic}
   \STATE {\bfseries Input:} Test points ${\bf x}_1, {\bf x}_2 \ldots, {\bf x}_t$, Noise vector $\bm{\epsilon} = (\epsilon, \ldots, \epsilon)$, redundancy budget $r$
   \STATE {\bfseries Output:} Optimized redundancy allocation vector ${\bf r}^r$
   \STATE {\bfseries Initialize} ${\bf r}(0)$ to 0
   \FOR{$j=0$ {\bfseries to} $r-1$}
   \STATE $i \leftarrow \text{arg}\min_{i} \frac{1}{t} \sum_{i=1}^t\text{SCP}_{app} ({\bf x_t}, {\bf E}^{{\bf r}(j)+{\bf 1}(i)}))$
   \STATE ${\bf r}(j+1) \leftarrow {\bf r}(j)+{\bf 1}(i)$
   \ENDFOR
\end{algorithmic}
\end{algorithm}

\begin{theorem}
Let $\bf x$ be a feature vector and $\bm{\epsilon} = (\epsilon_1, \ldots, \epsilon_{i-1},\epsilon_i, \epsilon_{i+1},\ldots, \epsilon_n)$ and $\bm{\epsilon}' = (\epsilon_1, \ldots, \epsilon_{i-1},\epsilon_i', \epsilon_{i+1},\ldots, \epsilon_n)$ be noise parameter vectors with $\epsilon_i \geq \epsilon'_i$ Let the associated error vectors be ${\bf E}, {\bf E}',$ respectively. If $D_i > 0$, then $\text{SCP}({\bf x}, {\bf E}) \geq \text{SCP}({\bf x}, {\bf E}')$, while if $D_i \leq 0$, then $\text{SCP}({\bf x}, {\bf E}) \leq \text{SCP}({\bf x}, {\bf E}')$.
\end{theorem}
\begin{IEEEproof}
Every summand in $\text{SCP}({\bf x}, {\bf E})$ is either a multiple of $\epsilon_i$ (if it corresponds to a subset of the $D_j$s that includes $D_i$) or a multiple of $(1-\epsilon_i)$ (if it corresponds to a subset of the $D_j$ that does not include $D_i$). We write \[\text{SCP}({\bf x}, {\bf E}) = \epsilon_i C_1 + (1-\epsilon_i) C_2,\]
where $C_1, C_2$ are non-negative.

\emph{Case 1)} $D_i > 0$. Consider the subsets of $\{D_1, \ldots, D_n\} \setminus \{D_i\}$. corresponding to $C_1$. These subsets must have sum larger than $T-D_i$, since including $D_i$, their sum must exceed $T$. In the case of $C_2$, these subsets have sum larger than $T$. Since $D_i$ is positive, any subset with sum larger than $T$ has sum larger than $T-D_i$ as well, so all subsets corresponding to $C_2$ also correspond to terms in $C_1$. We may write $C_1 = C_2 + C_3$ for some $C_3 \geq 0$. Then, 
\begin{align*}
\text{SCP}({\bf x}, {\bf E}) &= \epsilon_i C_1 + (1-\epsilon_i) C_2 \\
&= \epsilon_i(C_2+C_3) + (1-\epsilon_i) C_2 \\
&= C_2 + \epsilon_iC_3.
\end{align*}

Since $C_3 \geq 0$, reducing $\epsilon_i$ to $\epsilon'_i$ reduces the SCP as well.

\emph{Case 2)} $D_i \leq 0$. Then, $T-D_i \geq T$, so all subsets corresponding to $C_1$ also correspond to $C_2$. Thus, we can write 
\begin{align*}
\text{SCP}({\bf x}, {\bf E}) &= \epsilon_i C_1 + (1-\epsilon_i) C_2  \\
&= \epsilon_iC_1 + (1-\epsilon_i) (C_1+C_3) \\
&= C_1 + (1-\epsilon_i)C_3. 
\end{align*}
Reducing $\epsilon_i$ to $\epsilon'_i$ increases $(1-\epsilon_i)C_3$, so the SCP is increased.
\end{IEEEproof}

In words, noise helps the SCP when applied to those features that disagree with the classification, since flipping these bits increases confidence (and can cover up for other bit flips that reduce it). However, noise hurts the SCP when applied to features that agree with the classification; the sign of $D_i$ is a direct consequence of this idea.

The fact that reducing the error probability (by adding redundancy, etc.) is not always helpful leads us to seek an optimized solution. Clearly a uniform allocation of protection for all features is not always a good idea.

\begin{table*}
\centering
\caption{Experimental Results for Classification Change Probability (1-SCP)}
\label{tbl:ex}
\scalebox{1.0}{
\begin{tabular}{llll|lll|lll|lll}
\hline
\multicolumn{1}{l}{}    & \multicolumn{3}{l}{Movie dataset, feature set 1}                                        & \multicolumn{3}{l}{Movie dataset, feature set 2}                                        & \multicolumn{3}{l}{Voting dataset}                                                      & \multicolumn{3}{l}{NLTCS dataset}                                                         \\ \hline
\multicolumn{1}{l}{$R$} & \multicolumn{1}{l}{Uniform} & \multicolumn{1}{l}{Optimal} & \multicolumn{1}{l}{Ratio} & \multicolumn{1}{l}{Uniform} & \multicolumn{1}{l}{Optimal} & \multicolumn{1}{l}{Ratio} & \multicolumn{1}{l}{Uniform} & \multicolumn{1}{l}{Optimal} & \multicolumn{1}{l}{Ratio} & \multicolumn{1}{l}{$R_b$} & \multicolumn{1}{l}{Optimal} & \multicolumn{1}{l}{No protection} \\ \hline
1                         & 0.10112                      & 0.08664                      & 1.1670                     & 0.1263                       & 0.10245                      & 1.2328                     & 0.05635                      & 0.04135                      & 1.3628                     & 6                      & 0.0131                       & 0.0218                             \\
2                         & 0.03484                      & 0.03024                      & 1.1524                     & 0.04419                      & 0.03028                      & 1.4595                     & 0.03182                      & 0.02377                      & 1.3390                     & 7                      & 0.0120                       & 0.0218                             \\
3                         & 0.01154                      & 0.01002                      & 1.1513                     & 0.01469                      & 0.00879                      & 1.6711                     & 0.01903                      & 0.01429                      & 1.3319                     & 8                      & 0.0108                       & 0.0218                             \\
4                         & 0.00382                      & 0.00331                      & 1.1535                     & 0.00486                      & 0.00256                      & 1.9028                     & 0.01154                      & 0.0085                       & 1.3555                     & 9                      & 0.0097                       & 0.0218                             \\
5                         & 0.00127                      & 0.00109                      & 1.1576                     & 0.00162                      & 0.00077                      & 2.1019                     & 0.00702                      & 0.00504                      & 1.3911                     & 10                     & 0.0086                       & 0.0218                             \\
6                         & 0.00043                      & 0.00037                      & 1.1602                     & 0.00055                      & 0.00024                      & 2.2549                     & 0.00423                      & 0.00298                      & 1.4328                     & 11                     & 0.0074                       & 0.0218                             \\
7                         & 0.00015                      & 0.00013                      & 1.1622                     & 0.00019                      & 7.70E-05                     & 2.3983                     & 0.00261                      & 0.00177                      & 1.4742                     & 12                     & 0.0062                       & 0.0218                             \\
8                         & 4.94E-05                     & 4.24E-05                     & 1.1636                     & 6.30E-05                     & 2.48E-05                     & 2.539                      & 0.0016                       & 0.00105                      & 1.5182                     & 13                     & 0.0052                       & 0.0218                             \\
9                         & 1.69E-05                     & 1.45E-05                     & 1.1646                     & 2.16E-05                     & 8.08E-06                     & 2.6720                     & 0.00098                      & 0.00063                      & 1.5632                     & 14                     & 0.0044                       & 0.0218                             \\
10                        & 5.83E-06                     & 5.00E-06                     & 1.1654                     & 7.43E-06                     & 2.64E-06                     & 2.8109                     & 0.00060                      & 0.00038                      & 1.6107                     & 15                     & 0.0037                       & 0.0218\\ \hline                            
\end{tabular}}
\end{table*}

\subsection{Greedy Optimization}

In the remainder of this work, we introduce an algorithm to optimize the allocation of a redundancy budget for coded feature protection. We use a budget of $2r$ redundancy bits, which will be used to protect our $n$ features. Feature $i$ is then represented by $i+2r_i$ copies, and has error probability $\epsilon^{(r_i)}$, for $1  \leq i \leq n$. The noise vector for all $n$ features $\bm{\epsilon}^{({\bf r})} = (\epsilon^{(r_1)}, \epsilon^{(r_2)}, \ldots, \epsilon^{(r_n)})$. The corresponding error vector is written ${\bf E}^{(\bf r)}$. The values ${\bf r} = (r_1, r_2, \ldots, r_n)$ are constrained by $r_1 + r_2 + \ldots + r_n = r$. To minimize the distortion due to noise on point ${\bf x}$, we must maximize the SCP with respect to $\bf r$ and the resulting $\bm{\epsilon}$ vector. We have the following optimization
\begin{align*}
&\text{arg} \max_{\bf r} \text{SCP}({\bf x}, {\bf E}^{({\bf r})}) \quad \text{ s.t.} \sum_{i=1}^n r_i = r, \quad r_i \geq 0.
\end{align*}  

\begin{figure}
\centering
\includegraphics[width=.6\columnwidth]{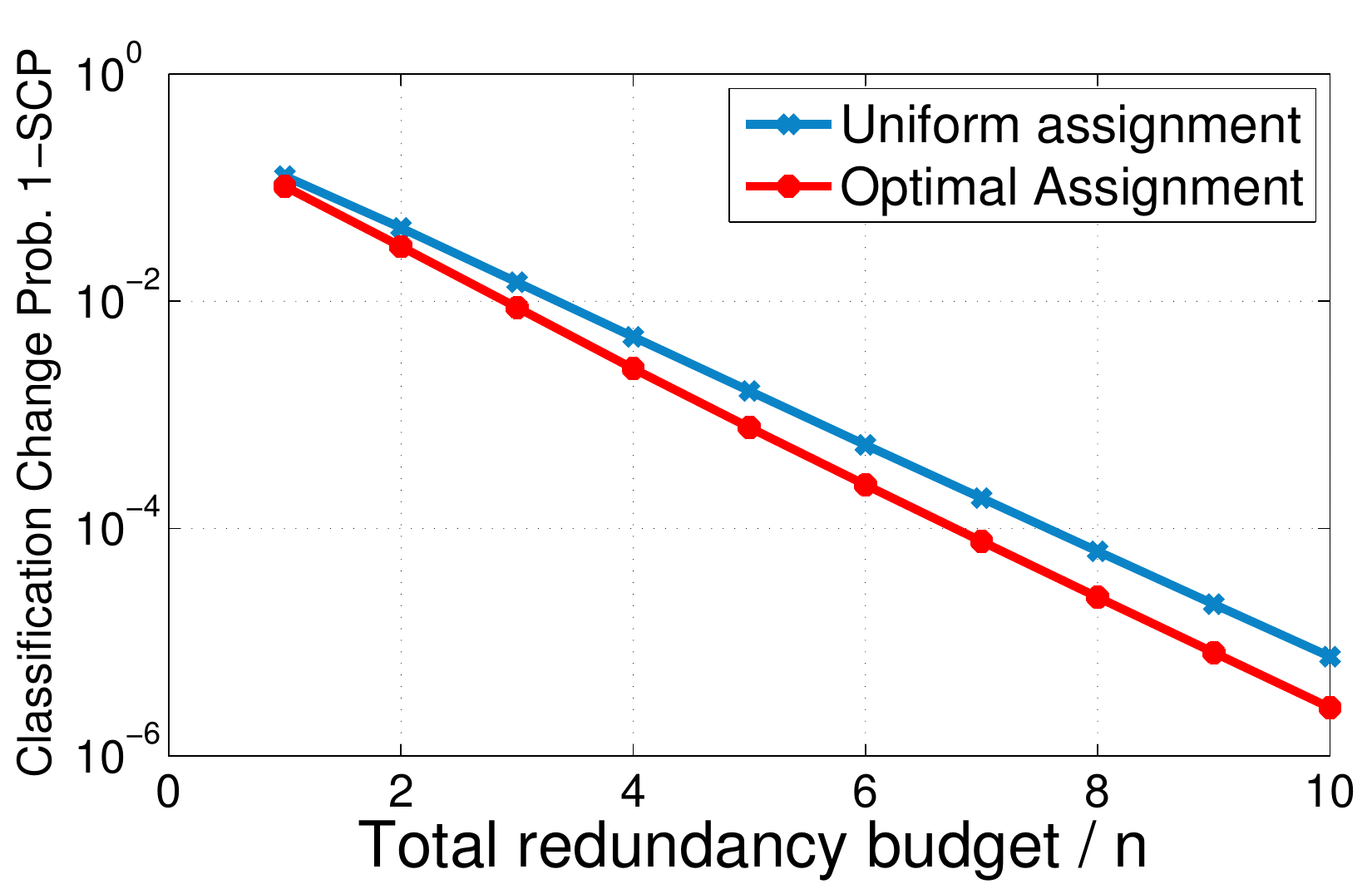}
\caption{Classification change probability ($1-\text{SCP}$) for a movie reviews data set with $n=20$ features and a test set of size of $t=250$. Comparison of the results of optimized versus uniform redundancy allocation.}
\label{fig:app3}
\end{figure}

Our first step is to perform a tractable version of this optimization using our SCP approximation:
\begin{align*}
&\text{arg} \max_{\bf r} \text{SCP}_{app}({\bf x}, {\bf E}^{({\bf r})}) \quad \text{ s.t.} \sum_{i=1}^n r_i = r, \quad r_i \geq 0.
\end{align*}  

Although we can now efficiently check each value of ${\bf r}$, doing so is still computationally expensive. The number of solutions to $r_1 + r_2 + \ldots + r_n= r$ with $r_i \geq 0$ is given by $\binom{n+r-1}{r}$. We turn to a greedy (myopic) optimization approach to further reduce the complexity. This approach enables us to perform the optimization one redundancy unit (two repeated bits) at a time. After the $j$th step, we write the redundancy vector as ${\bf r}(j)  = (r_1(j), r_2(j), \ldots, r_n(j))$. In the $(j+1)$st step, one of the $r_i(j)$ terms is selected and increased by 1. For ease of notation, let us write ${\bf 1}(i)$ for the vector $(0,0,\ldots,0,1,0,\ldots,0)$ with a $1$ in the $i$th position and $0$s elsewhere. Then, the $(j+1)$st step is given by 
\begin{align*}
&\text{arg}\max_{i} \text{SCP}_{app}({\bf x}, {\bf E}^{{\bf r}(j)+{\bf 1}(i)})) \quad \text{ s.t. } 1 \leq i \leq n.
\end{align*}  

This reduces our complexity to that of performing $n$ SCP computations per each of the $r$ steps. Of course, we can also perform the optimization over $t$ test data points. The procedure is given in Algorithm~\ref{alg:code} 


\begin{figure}
\centering
\includegraphics[width=.6\columnwidth]{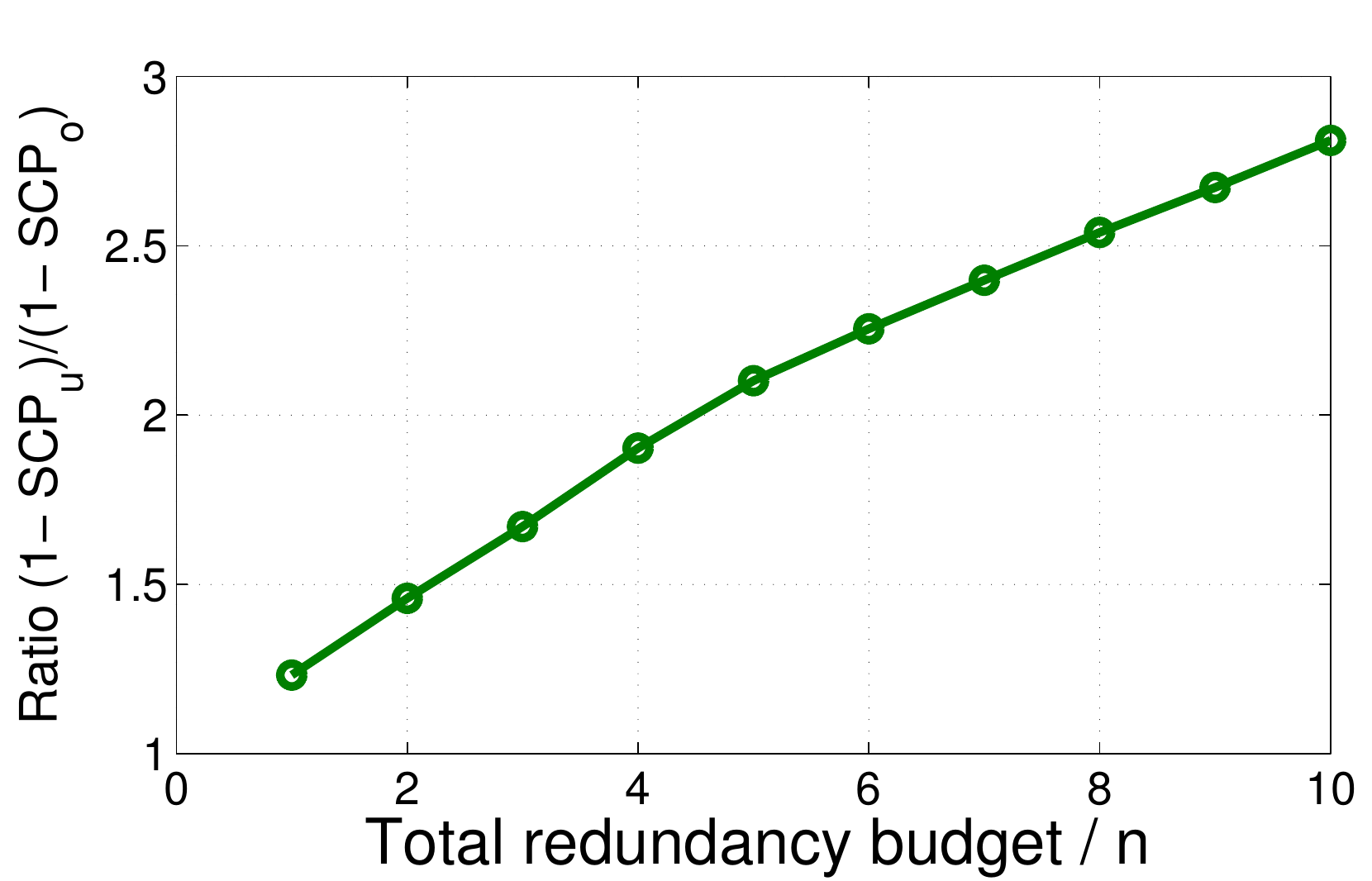}
\caption{The classification change probability (uniform to optimized allocation) ratio reaches $3\times$. The setup is the same as that of Figure~\ref{fig:app}.}
\label{fig:app4}
\end{figure}

We demonstrate our approach with SCP experiments using the voting dataset \cite{Lichman}, the movie review dataset from \cite{Bekker}, and the NLTCS dataset in \cite{Lowd}. In all cases, we used $k=50$ for the number of buckets in our SCP approximations. For the movie reviews set we used two sets of $n=20$ features and $250$ test data points. The $\epsilon$ parameter was $0.2$ for the first 2 datasets (a very high amount of noise that can nevertheless be handled through channel coding). Table~\ref{tbl:ex} shows the classification change ($1-\text{SCP}$) probability for uniform and optimized assignment of redundancy bits given a budget of $2nR$ bits. We report the ratio of the classification change probability (uniform versus optimized); this ratio reaches up to $3\times$. 

In the movie review dataset, the first feature set included features with similar dependence on the class variable. The second feature set included a mixture of features (some more informative of the class variable than others). In this case the optimized assignment has a clear benefit over the uniform assignment. The results on the voting dataset (using $n=16$ features and $t=454$ test data points) are in between those of the two movie review sets. A plot of the SCP and the improvement ratio for the more beneficial assignment is shown in Figures~\ref{fig:app3} and ~\ref{fig:app4}.

For the NLTCS dataset, we tested the case of a small amount of redundancy $R_b \leq 15$, so that a uniform allocation is impossible, and we must rely on the optimization. Here, $n=15$ features and $t=3236$ test data points were used. The results are compared to the unprotected version.


\section{Conclusion}
We studied binary na\"{\i}ve Bayes classifiers operating on noisy test data. First, we characterized the impact of bit flips due to noise on the classifier output with the same classification probability (SCP). We introduced a low-complexity approximation for the SCP based on quantization and polynomial multiplication. Next, we considered minimizing the classifier distortion by allocating redundant bits among the features. Our informed approached for redundancy allocation is among the first principled methods combining coding theory with machine learning. We demonstrated the results of this idea with experiments.

\bibliography{bib_draft}

\begin{thebibliography}{10}

\bibitem{Shannon}
C.~E. Shannon, ``Communication in the presence of noise,'' {\em Proceedings of
  the IEEE}, vol.~86, no.~2, pp.~447--457, 1998.

\bibitem{Dilillo}
L.~Dilillo, P.~Girard, S.~Pravossoudovitch, A.~Virazel, and M.~B. Hage-Hassan,
  ``Data retention fault in {SRAM} memories: analysis and detection
  procedures,'' in {\em Proceedings of the 23rd IEEE VLSI Test Symposium
  (VTS)}, 2005.

\bibitem{Mittal}
S.~Mittal, ``A survey of techniques for approximate computing,'' {\em ACM
  Computing Surveys}, vol.~48, no.~4, 2016.

\bibitem{Chandra}
V.~Chandra and R.~Aitken, ``Impact of technology and voltage scaling on the
  soft error susceptibility in nanoscale {CMOS},'' in {\em Proc. Defect and
  Fault Tolerance of VLSI Systems (DFTVS `08)}, (Boston, MA), 2008.

\bibitem{Sridharan}
V.~Sridharan and D.~Liberty, ``A field study of {DRAM} errors,'' in {\em Proc.
  International Conference for High Performance Computing, Networking, Storage
  and Analysis (SC)}, (Salt Lake City, UT), 2012.

\bibitem{Riggle}
C.~M. Riggle and S.~G. McCarthy, ``Design of error correction systems for disk
  drives,'' {\em IEEE Transactions on Magnetics}, vol.~34, no.~4,
  pp.~2362--2371, 1998.

\bibitem{Rivest}
R.~L. Rivest and A.~Shamir, ``How to reuse a ``write-once memory'','' {\em
  Information and Control}, vol.~55, no.~1-3, pp.~1--19, 1982.

\bibitem{Blaum}
M.~Blaum, J.~Brady, J.~Bruck, and J.~Menon, ``{EVENODD}: an efficient scheme
  for tolerating double disk failures in {RAID} architectures,'' {\em IEEE
  Transactions on Computers}, vol.~44, no.~2, pp.~192--202, 1995.

\bibitem{Cassuto}
Y.~Cassuto, M.~Schwartz, V.~Bohossian, and J.~Bruck, ``Codes for asymmetric
  limited-magnitude errors with application to multilevel flash memories,''
  {\em IEEE Transactions on Information Theory}, vol.~56, no.~4,
  pp.~1582--1595, 2010.

\bibitem{Dolecek}
L.~Dolecek and F.~Sala, ``Channel coding methods for non-volatile memories,''
  {\em Foundations and Trends in Communications and Information Theory},
  vol.~13, no.~1, pp.~1--136, 2016.

\bibitem{Weatherspoon}
H.~Weatherspoon and J.~Kubiatowicz, ``Erasure coding vs. replication: A
  quantitative comparison,'' in {\em Proceedings of First International
  Workshop on Peer-to-Peer Systems (IPTPS '01)}, 2002.

\bibitem{Dimakis}
A.~G. Dimakis, P.~B. Godfrey, Y.~Wu, M.~J. Wainwright, and K.~Ramchandran,
  ``Network coding for distributed storage systems,'' {\em IEEE Transactions on
  Information Theory}, vol.~56, no.~9, pp.~4539--4551, 2010.

\bibitem{KZhao}
K.~Zhao, W.~Zhao, H.~Sun, T.~Zhang, X.~Zhang, and N.~Zheng, ``{LDPC}-in-{SSD}:
  Making advanced error correction codes work effectively in solid state
  drives,'' in {\em Proceedings of Conference on File and Storage Technologies
  (FAST `13)}, (San Jose, CA), 2013.

\bibitem{Mazooji16}
K.~Mazooji, F.~Sala, G.~{Van den Broeck}, and L.~Dolecek, ``Robust channel
  coding strategies for machine learning data,'' in {\em Proc. Annual Allerton
  Conference on Communication, Control, and Computing (Allerton)},
  pp.~609--616, 2016.

\bibitem{Srivastava}
N.~Srivastava, G.~E. Hinton, A.~Krizhevsky, I.~Sutskever, and R.~Salakhutdinov,
  ``Dropout: a simple way to prevent neural networks from overfitting,'' {\em
  Journal of Machine Learning Research}, vol.~15, no.~1, pp.~1929--1958, 2014.

\bibitem{Achille}
A.~Achille and S.~Soatto, ``Information dropout: Learning optimal
  representations through noisy computation,'' tech. rep., University of
  California, Los Angeles, 2016.

\bibitem{Kala}
E.~Kalapanidas, N.~Avouris, M.~Craciun, and D.~Neagu, ``Machine learning
  algorithms: a study on noise sensitivity,'' in {\em Proc. 1st Balcan
  Conference in Informatics}, (Thessaloniki, Greece), pp.~356--365, 2003.

\bibitem{Dekel}
O.~Dekel and O.~Shamir, ``Learning to classify with missing and corrupted
  features,'' in {\em Proc. 25th International Conference on Machine Learning
  (ICML 2008)}, (Helsinki, Finland), 2008.

\bibitem{Globe}
A.~Globerson and S.~Roweis, ``Nightmare at test time: robust learning by
  feature deletion,'' in {\em Proc. 25th International Conference on Machine
  Learning (ICML 2006)}, (Pittsburgh, PA), 2006.

\bibitem{Ramoni}
M.~Ramoni and P.~Sebastiani, ``Robust bayes classifiers,'' {\em Artificial
  Intelligence}, vol.~125, no.~1-2, pp.~209--226, 2001.

\bibitem{Zhang}
J.~C. Duchi, M.~I. Jordan, M.~J. Wainwright, and Y.~Zhang, ``Optimality
  guarantees for distributed statistical estimation,'' tech. rep., University
  of California, Berkeley, 2014.

\bibitem{Han}
T.~S. Han and S.~Amari, ``Statistical inference under multiterminal data
  compression,'' {\em IEEE Transactions on Information Theory}, vol.~44, no.~6,
  pp.~2300--2324, 1998.

\bibitem{Heckerman93}
D.~Heckerman, E.~Horvitz, and B.~Middleton, ``An approximate nonmyopic
  computation for value of information,'' {\em IEEE Transactions on Pattern
  Analysis and Machine Intelligence}, vol.~15, no.~3, pp.~292--298, 1993.

\bibitem{bilgic11}
M.~Bilgic and L.~Getoor, ``Value of information lattice: Exploiting
  probabilistic independence for effective feature subset acquisition,'' {\em
  Journal of Artificial Intelligence Research (JAIR)}, vol.~41, pp.~69--95,
  2011.

\bibitem{krause2009}
A.~Krause and C.~Guestrin, ``Optimal value of information in graphical
  models,'' {\em Journal of Artificial Intelligence Research (JAIR)}, vol.~35,
  pp.~557--591, 2009.

\bibitem{darwiche12}
A.~Choi, Y.~Xue, and A.~Darwiche, ``{S}ame-{D}ecision {P}robability: A
  confidence measure for threshold-based decisions,'' {\em International
  Journal of Approximate Reasoning (IJAR)}, vol.~2, pp.~1415--1428, 2012.

\bibitem{chen2014}
S.~Chen, A.~Choi, and A.~Darwiche, ``Algorithms and applications for the
  {S}ame-{D}ecision {P}robability,'' {\em Journal of Artificial Intelligence
  Research (JAIR)}, vol.~49, pp.~601--633, 2014.

\bibitem{chen2015computer}
S.~Chen, A.~Choi, and A.~Darwiche, ``Computer adaptive testing using the
  same-decision probability,'' in {\em Proceedings of the Twelfth UAI
  Conference on Bayesian Modeling Applications Workshop}, pp.~34--43, 2015.

\bibitem{gimenez2014novel}
F.~J. Gimenez, Y.~Wu, E.~S. Burnside, and D.~L. Rubin, ``A novel method to
  assess incompleteness of mammography reports,'' in {\em AMIA Annual Symposium
  Proceedings}, vol.~2014, p.~1758, American Medical Informatics Association,
  2014.

\bibitem{Frenay}
B.~Frenay and A.~Kaban, ``A comprehensive introduction to label noise,'' in
  {\em Proc. European Symposium on Artificial Neural Networks, Comp.
  Intelligence and Machine Learning (ESANN)}, (Bruges, Belgium), 2014.

\bibitem{Vempaty}
A.~Vempaty, L.~R. Varshney, and P.~K. Varshney, ``Reliable crowdsourcing for
  multi-class labeling using coding theory,'' {\em IEEE Journal of Selected
  Topics in Signal Processing}, vol.~8, no.~4, pp.~667--679, 2014.

\bibitem{Guyon}
I.~Guyon and A.~Elisseeff, ``An introduction to variable and feature
  selection,'' {\em Journal of Machine Learning Research}, vol.~3,
  pp.~1157--1182, 2003.

\bibitem{gfref}
H.~S. Wilf, {\em Generatingfunctionology}.
\newblock A K Peters/CRC Press, 3rd~ed., 2005.

\bibitem{fftref}
K.~R. Rao, D.~N. Kim, and J.-J. Hwang, {\em Fast Fourier Transform - Algorithms
  and Applications}.
\newblock Springer Publishing Company, 1st~ed., 2010.

\bibitem{Lichman}
M.~Lichman, ``{UCI} machine learning repository,'' 2013.

\bibitem{Bekker}
J.~Bekker, J.~Davis, A.~Choi, A.~Darwiche, and G.~{Van den Broeck}, ``Tractable
  learning for complex probability queries,'' in {\em Proceedings of Conference
  on Neural Information Processing Systems (NIPS)}, (Montreal, Canada), 2015.

\bibitem{Lowd}
D.~Lowd and J.~Davis, ``Learning markov network structure with decision
  trees,'' in {\em Proc. IEEE International Conference on Data Mining (ICDM)},
  (Sydney, Australia), 2010.

\end{thebibliography}

\end{document}